\documentclass[final]{cvpr}
\usepackage{booktabs}
\usepackage{times}
\usepackage{epsfig}
\usepackage{cite}
\usepackage{graphicx}
\usepackage{amsmath}
\usepackage{amssymb}
\usepackage{multirow}

\usepackage[pagebackref=true,breaklinks=true,colorlinks,bookmarks=false]{hyperref}

\def\cvprPaperID{****} 
\def\confYear{CVPR 2021}

\begin{document}

\title{2rd Place Solutions in the HC-STVG track of Person in Context Challenge 2021}

\author{
YiYu
\and 
XinyingWang
\and 
WeiHu
\and 
XunLuo
\and 
ChengLi
\and
\\
\tt MGTV
\\
{\tt\small\{yuyi, xinying, huwei, luoxun, licheng\}@mgtv.com}
}

\maketitle

\begin{abstract}
In this technical report, we present our solution to localize a spatio-temporal person in an untrimmed video based on a sentence.
We achieve the second vIOU(0.30025) in the HC-STVG track of the 3rd Person in Context(PIC) Challenge. Our solution contains three parts: 1) human attributes information is extracted from the sentence, it is helpful to filter out tube proposals in the testing phase and supervise our classifier to learn appearance information in the training phase. 2) we detect humans with YoloV5 and track humans based on the DeepSort framework but replace the original ReID network with FastReID. 3) a visual transformer is used to extract cross-modal representations for localizing a spatio-temporal tube of the target person.
\end{abstract}

\section{Dataset}
The Human-centric Spatio-Temporal Video Grounding (HC-STVG) \cite{tang2020human} dataset provides 16k annotation-video pairs with different movie scenes. The duration of each video is 20 seconds. The video includes multiple people and the trajectories of the corresponding person are annotated. The task of the challenge is to output the start frame and the end frame number with the bounding boxes of the target person during the video clip given a description. There are about 12k videos adopted for training and validation data and 4.4k videos for testing data.

\section{Methodology}

\begin{figure*}
\centering
 \includegraphics[width=\textwidth]{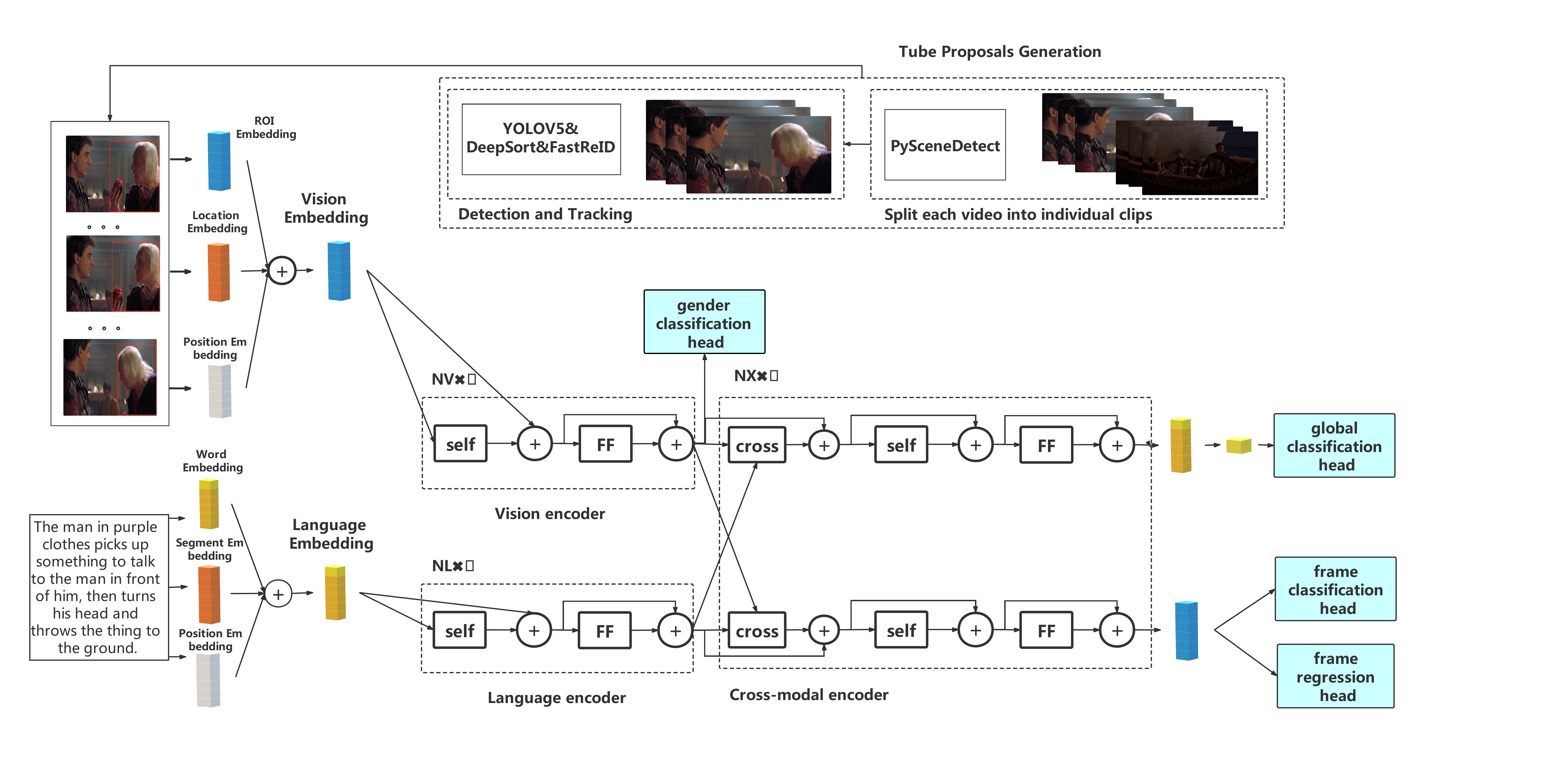}
 \caption{\label{fig:network}The overall pipeline of our method. }
\end{figure*}

Our method contains the following steps:  1) It first analyses video description using NLP technology to extract human information such as gender, clothing color and clothing type. 2) Secondly, the video is split into separate clips by detecting scene changes. After that, to generate the tube proposals, we detect humans per frame and link them within the same clip. 3) Thirdly, a pretrained Mask R-CNN~\cite{2017Mask} model will be used to extract bounding box feature, and then a cross-modal transformer takes visual feature and corresponding sentence as inputs to extract cross-modal feature and output global classification score, gender classification score, frame classification score and frame regression score.

\subsection{Triple extraction description}

We use the NLP technology\cite{2009TFIDF,2014EMNLP} containing regular matching, keyword extraction, syntactic analysis to extract the human appearance attributes from the description. There are triples of text content extracted from each sentence. For example, below are the sentences and corresponding human appearance attributes:\\
``The woman in the green dress walks to the woman in the red dress.''\\
tuples: ((female, green, cloth), (female, red, cloth))\\
Basically, we divided the characters into male and female categories to quickly distinguish human gender. And according to the contrast of the colors, we divide the clothing colors into 11 categories(white, black, gray, red, orange, yellow, green, blue, purple, pink, brown) and divide the clothing type into three categories(top, bottom, cloth). \\
These tuples indicate the number of people who appeared in this description and the appearance of the main subject. The number of people is helpful to select the scene and reduce the tube candidates. The appearance features are used later to help our classification model and supervise our network to learn the human appearance information.

\subsection{Tube proposals generation}
Given an video, we firstly determine the exact frames where scene transitions occur and split each video into individual clips by using an open-source algorithm called PySceneDetect. Secondly, we detect human bounding boxes in each frame by using YoloV5 with the NMS confidence threshold equals  0.4 and the NMS IoU threshold equals 0.65. Next, for each video clip, we follow the DeepSort\cite{Wojke2017simple,Wojke2018deep} framework to connect human bounding boxes, additionally, we change the original ReID network to FastReID\cite{he2020fastreid} for better performance. After that, we delete the tube proposal which has a large time overlap and a large bounding box IoU with another tube. Also, within the same clip, we reconnect the tube proposals which simultaneously satisfy the following two conditions: (1) one tube towards the end and another is about to begin, (2) they have large IoU during their overlapping time. At Last, a simple Gaussian filter is applied to smooth tube tracks.
In the testing phase, we ignore the video clip which only contains one tube while the description contains more than one character. 

\subsection{Cross-modal representation}
This task involves cross-modality features: visual information and textural description. We adopt a cross-modal transformer to exploit the relationship between these two modalities. Following\cite{tan2019lxmert}, we first construct an embedding layer to convert the input(i.e. a tube proposal and a sentence) into two sequences of features: frame-level bounding box embeddings and word-level sentence embeddings. After the embedding layer, we apply two Single-Modality Encoders, i.e., a language encoder and a vision encoder, and each of them only focuses on a single modality. It is worth noting that, We add a gender classification branch at the end of the vision encoder to supervise our visual network learning gender information. Finally, we use a Cross-Modality Encoder which will output two sequence features $f^{frame}_{p,t}$ and $f^{word}_{p,t}$. $f^{word}_{p,t}$ is produced by using the textual inputs as the Query and the visual inputs as the Key and Value and $f^{frame}_{p,t}$ is the opposite. We take the first token position feature of $f^{word}_{p,t}$ as the global feature,i.e. $f^{global}_{p}$, to predict whether the proposal tube matches the query sentence. The $f^{frame}_{p,t}$ is used to do tube trimming prediction.

\subsection{Tube-description Matching}
In this step, to prepare the positive and negative samples, we verify which tube proposal $T_{p}$ is matching with the query description $s$. The tube proposals which simultaneously satisfy the following two scores are treated as positive samples: $S_{overlap}>0.7$ and $S_{IoU}>0.5$. While the tube proposals are treated as negative samples when $S_{overlap} <0.3$ or $S_{IoU}<0.3$. $S_{overlap}$ and $S_{IoU}$ are defined as followed:
\begin{equation}
  S_{overlap} =\frac{\left | T_{p}\cap T_{GT} \right |}{\left | T_{GT}\right |} 
\end{equation}
\begin{equation}
  S_{IoU}={\frac{1}{ \left | T_{p}\cap T_{GT} \right |}}\sum_{t\in T_{p}\cap T_{GT}} IOU(B_{t}^{'},B_{t})  
\end{equation}
where $T_{p}$ and $T_{GT}$ are the set of frames of selected and ground truth tube. $B_{t}^{'}$ and $B_{t}$ are predicted bounding box and ground truth bounding box of frame $t$. 

\subsection{Tube Trimming}
Tube trimming is to localize the temporal boundary of the target person since tube proposals may contain redundant transition frames. We follow \cite{tang2020human} to apply tube trimming. During the experiment, we find that the performance with tube trimming is even worse. But if training without tube trimming branch loss will harm the global classification accuracy. So we treat the tube trimming branch as an auxiliary loss and this branch does not participate in the inference process.
\subsection{Loss Functions}
For a tube-sentence pair $T_p$, the model will output global classification score $GL^{'}_p$, gender classification score $GE^{'}_p$,frame-level classification score $C^{'}_{p,t}$ and frame-level regression score $R^{'}_{p,t}$. $GL_p$,$GE_p$, $C_{p,t}$ and $R_{p,t}$ are corresponding ground-truth labels. the total loss is defined as follows,
\begin{equation}
\begin{split}
    L=\sum_{p} \left \{ \right . (\lambda_1*L_{focal}(GL^{'}_p, GL_p) \\+
    \lambda_2*\mathbb{I}_{\left \{ GL_p=1 \right \}}L_{gender}(GE^{'}_p, GE_p)\\+
    \lambda_3*\mathbb{I}_{\left \{ GL_p=1 \right \}} \frac{1}{N}\sum_{t}L_{cls}(C^{'}_{p,t}, C_{p,t}) 
    \\  + \lambda_4*\mathbb{I}_{\left \{ GL_p=1,C_{p,t}=1 \right \} }\frac{1}{N_{pos}}\sum_{t}L_{reg}(R^{'}_{p,t}, R_{p,t}) \left. \right \}
\end{split}
\end{equation}
where $L_{focal}$ is the focal loss\cite{2017Focal} for tube-sentence matching classification, $L_{cls}$ and $L_{gender}$ are both the cross-entropy loss and $L_{reg}$ is the  IoU loss (i.e, $-ln(\frac{R^{'}_{p,t}\cap R_{p,t}}{R^{'}_{p,t}\cup R_{p,t}})$) for frame classification, gender classification only base on vision information and frame regression, respectively. $N$ and $N_{pos}$ is number of frames and positive frames in the tube $T_p$.

\subsection{Evaluation Metric}
The challenge takes $vIoU$ as the final evaluation metric, it can directly reflect the accuracy of the prediction results spatiotemporally.
\begin{equation}
  vIoU={\frac{1}{\left | S_{u}\right | }}\sum_{t\in S_{i}} IOU(B_{t}^{'},B_{t})  
\end{equation}
where $S_{i}$ is the set of frames in the intersection of selected and ground truth tube, $S_{u}$ is the set of frames in the union of selected and ground truth tube, $B_{t}^{'}$ and $B_{t}$ are predicted bounding box and ground truth bounding box of frame $t$.

\section{Experiment}
In this section, we will introduce how we improve the baseline model step by step. 
\subsection{Ablation Study}
After analyzing the result of our baseline model, we find three potential improvements: 1) We observe the positive and negative tubes are heavily imbalanced. To solve this issue, we replace the cross-entropy loss with the focal loss in global classification. 2) Our baseline model is over-fitting in the training dataset. For this problem, we take two solutions: one is to freeze the language encoder parameter during training and another is to apply data augmentation in the tube level, however the data augmentation does not contribute a lot. 3) As we analyzing some failure cases, we find that our baseline model pays more attention to distinguish the human behaviors but ignores some obvious character features like human appearance. Thus, we add a gender classification branch at the end of our vision encoder. \\
The results of our ablation experiments are summarized in Table~\ref{tab:ablation}. In this table, +FO means applying the focal loss in the global classification network. +FI and +GE mean freezing language encoder parameter and applying gender loss during training phase respectively.

\begin{table}[ht]
    \centering
    \caption{ Ablation Study}
    \begin{tabular}{c|c}
    \toprule
        Methods & vIoU \\
        \midrule
        baseline & 0.2775 \\
        baseline+FO & 0.2880 \\
        baseline+FO+FI & 0.2906 \\
        baseline+FO+FI+GE & 0.3002\\
    \bottomrule
    \end{tabular}
    \label{tab:ablation}
\end{table}

{\small
\bibliographystyle{unsrt}
\bibliography{egbib}

\begin{thebibliography}{1}

\bibitem{tang2020human}
Zongheng Tang, Yue Liao, Si~Liu, Guanbin Li, Xiaojie Jin, Hongxu Jiang, Qian
  Yu, and Dong Xu.
\newblock Human-centric spatio-temporal video grounding with visual
  transformers.
\newblock {\em arXiv preprint arXiv:2011.05049}, 2020.

\bibitem{2017Mask}
K.~He, G.~Gkioxari, P~Dollár, and R.~Girshick.
\newblock Mask r-cnn.
\newblock In {\em IEEE Transactions on Pattern Analysis \& Machine
  Intelligence}, 2017.

\bibitem{2009TFIDF}
C.~Y. Shi, X.~U. Chao-Jun, and X.~J. Yang.
\newblock Study of tfidf algorithm.
\newblock {\em Journal of Computer Applications}, 2009.

\bibitem{2014EMNLP}
D.~Chen and C.~Manning.
\newblock A fast and accurate dependency parser using neural networks.
\newblock In {\em Proceedings of the 2014 Conference on Empirical Methods in
  Natural Language Processing (EMNLP)}, 2014.

\bibitem{Wojke2017simple}
Nicolai Wojke, Alex Bewley, and Dietrich Paulus.
\newblock Simple online and realtime tracking with a deep association metric.
\newblock In {\em 2017 IEEE International Conference on Image Processing
  (ICIP)}, pages 3645--3649. IEEE, 2017.

\bibitem{Wojke2018deep}
Nicolai Wojke and Alex Bewley.
\newblock Deep cosine metric learning for person re-identification.
\newblock In {\em 2018 IEEE Winter Conference on Applications of Computer
  Vision (WACV)}, pages 748--756. IEEE, 2018.

\bibitem{he2020fastreid}
Lingxiao He, Xingyu Liao, Wu~Liu, Xinchen Liu, Peng Cheng, and Tao Mei.
\newblock Fastreid: A pytorch toolbox for general instance re-identification.
\newblock {\em arXiv preprint arXiv:2006.02631}, 2020.

\bibitem{2019LXMERT}
H.~Tan and M.~Bansal.
\newblock Lxmert: Learning cross-modality encoder representations from
  transformers.
\newblock 2019.

\bibitem{2017Focal}
T.~Y. Lin, P.~Goyal, R.~Girshick, K.~He, and P~Dollár.
\newblock Focal loss for dense object detection.
\newblock In {\em IEEE Transactions on Pattern Analysis \& Machine
  Intelligence}, pages 2999--3007, 2017.

\end{thebibliography}
}

\end{document}